%% file: main.tex
\documentclass[10pt,twocolumn,letterpaper]{article}

\usepackage[pagenumbers]{cvpr} 

\input{preamble}
\definecolor{cvprblue}{rgb}{0.21,0.49,0.74}
\usepackage[pagebackref,breaklinks,colorlinks,allcolors=cvprblue]{hyperref}
\usepackage[utf8]{inputenc} 
\usepackage[T1]{fontenc}    
\usepackage{hyperref}       
\usepackage{url}            
\usepackage{booktabs}       
\usepackage{amsfonts}       
\usepackage{nicefrac}       
\usepackage{microtype}      
\usepackage[dvipsnames]{xcolor}        
\usepackage{graphicx}
\usepackage{subcaption}  
\usepackage{caption}     
\usepackage{amsmath}
\usepackage{multirow}
\usepackage{booktabs}
\usepackage{pifont}
\usepackage{makecell}

\newcommand\blfootnote[1]{%
\begingroup
\renewcommand\thefootnote{}\footnote{#1}%
\addtocounter{footnote}{-1}%
\endgroup
}

\def\confName{CVPR}
\def\confYear{2026}

\title{Reading or Reasoning? Format Decoupled Reinforcement Learning for\\ Document OCR}

\author{Yufeng Zhong$^{*}$ \quad Lei Chen$^{*}$ \quad Zhixiong Zeng$^{\dagger}$ \quad Xuanle Zhao \quad Deyang Jiang \quad Liming Zheng \\
Jing Huang \quad Haibo Qiu \quad Peng Shi \quad Siqi Yang \quad Lin Ma$^{\ddagger}$ \\ \\
{Meituan}\\
Emails: <zengzhixiong@meituan.com, forest.linma@gmail.com> \\
Project: \href{https://github.com/DocTron-hub/FD-RL}{https://github.com/DocTron-hub/FD-RL}
}

\begin{document}

\maketitle

\blfootnote{$*$ Equal contribution. $\dagger$ Project leader. $\ddagger$ Corresponding author.}

\input{sec/0_abstract}
\input{sec/1_intro}
\input{sec/2_related_work}
\input{sec/3_method}

\input{sec/4_experiments}
\input{sec/5_conclusion}
{
    \small
    \bibliographystyle{ieeenat_fullname}
    \bibliography{main}
}

\input{sec/X_suppl}

\end{document}

%% file: sec/0_abstract.tex
\begin{abstract}

Reading text from images or scanned documents via OCR models has been a longstanding focus of researchers. Intuitively, text reading is perceived as a straightforward perceptual task, and existing work primarily focuses on constructing enriched data engineering to enhance SFT capabilities. In this work, we observe that even advanced OCR models exhibit significantly higher entropy in formatted text (\emph{e.g.}, formula, table, etc.) compared to plain text, often by an order of magnitude. These statistical patterns reveal that advanced OCR models struggle with high output uncertainty when dealing with format sensitive document, suggesting that reasoning over diverse reading pathways may improve OCR performance.
To address this, we propose format decoupled reinforcement learning (FD-RL), which leverages high-entropy patterns for targeted optimization. Our approach employs entropy-based data filtration strategy to identify format-intensive instances, and adopt format decoupled rewards tailored to different format types, enabling format-level validation rather than token-level memorization. 
FD-RL achieves an average score of 90.41 on OmniDocBench, setting a new record for end-to-end models on this highly popular benchmark.
More importantly, we conduct comprehensive ablation studies over data, training, filtering, and rewarding strategies, thoroughly validating their effectiveness. 

\end{abstract}

%% file: sec/1_intro.tex
\section{Introduction}
\label{sec:intro}
Document parsing via Optical Character Recognition (OCR) aims to recognize and extract text from document images, which plays a vital role in diverse application domains. Given its importance, numerous advanced methods have recently been proposed, such as Deepseek-OCR~\cite{wei2025deepseek}, dots.ocr~\cite{rednote2025dotsocr}, and PaddleOCR-VL~\cite{cui2025paddleocr}.



\begin{figure}[!t]
\centering
\includegraphics[width=0.95\linewidth]{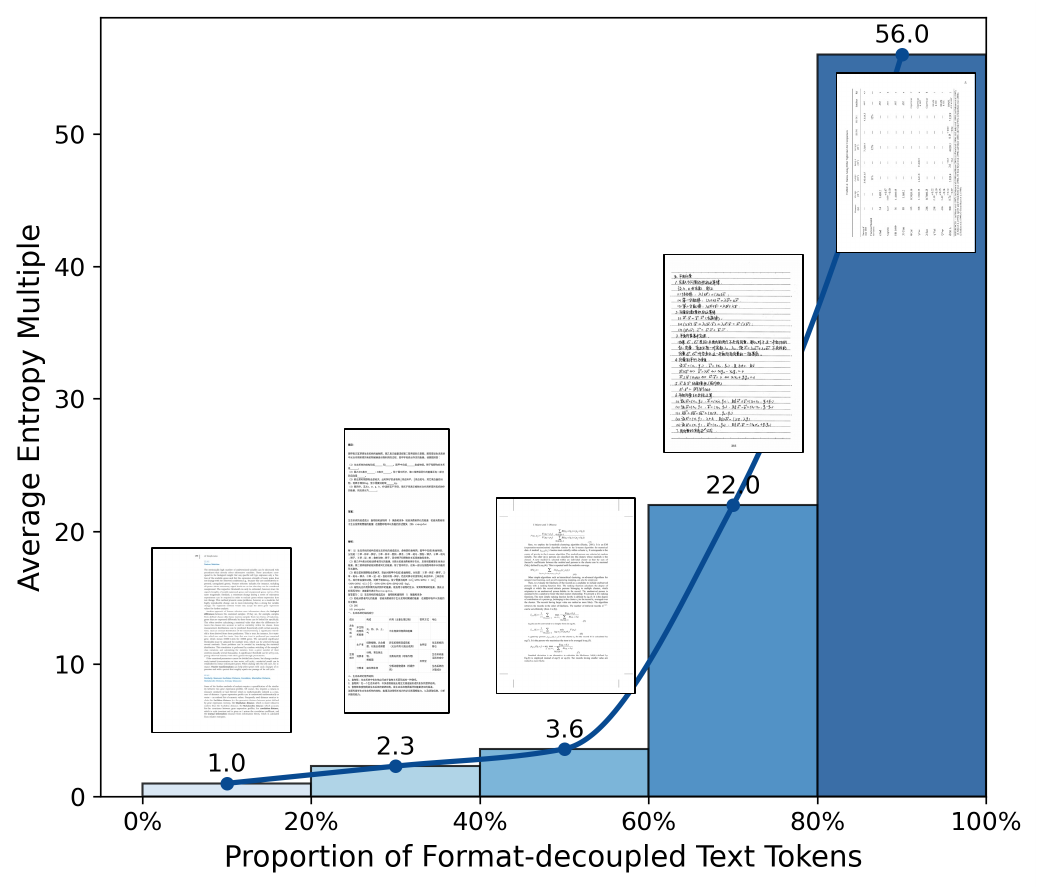}
\caption{Document samples with a high proportion of formatted text exhibit high entropy properties. We divide the document content into \textbf{plain text} and \textbf{formatted text}. The horizontal axis represents the proportion of formatted text to the total document content, and the vertical axis represents the ratio of token entropy.
We present a representative document example for each type.}
\label{fig:entropy_vs_pure_text}
\end{figure}

Traditional document OCR methods ~\cite{wang2024mineru,cui2025paddleocr,Paruchuri2025Marker}  utilize a two-stage pipeline that first detects layout information and then employs specialized parsers to recognize text. While these pipeline-based methods offer high stability and performance, they suffer from limited flexibility and high fine-tuning costs. In contrast, vision-language model (VLM)-based methods~\cite{ocrflux2025,mistral2025ocr,liu2025points,poznanski2025olmocr-1,mandalm2025nanonets,wei2025deepseek,rednote2025dotsocr} employ VLMs to decode text directly from visual inputs, showing simplified workflows and superior generalization capabilities. Recently, some pipeline-based VLM methods~\cite{feng2025dolphin,li2025monkeyocr,niu2025mineru2,cui2025paddleocr-vl} integrate the advantages of both approaches, significantly improving the accuracy of VLM-based methods by utilizing pre-detected layout information.

Although existing work achieves strong performance through meticulously designed data engineering and training pipelines, they still exhibit significantly higher entropy in formatted text (\emph{e.g.}, formula, table, etc) compared to plain text tokens. 
As shown in Figure~\ref{fig:entropy_vs_pure_text}, we conduct an empirical analysis across a large corpus, categorizing document samples by their formatted text proportion into five bins: 20\%, 40\%, 60\%, 80\%, and 100\%. We then compute the average output entropy during inference based on Qwen3-VL~\cite{Qwen3-VL} for all samples within each bin.
The empirical results reveal a clear pattern: document samples with higher formatted text proportions exhibit substantially higher entropy, indicating lower output certainty and larger model confusion.

In fact, existing work shows that the high entropy tokens act as critical forks to steer the language model toward diverse reasoning pathways~\cite{wang2025beyond}.
Therefore, the formatted text with high entropy may generate diverse reading pathways, producing diverse reward information that can facilitate reinforcement learning algorithms such as GRPO ~\cite{shao2024deepseekmath}.
A simple example is that formulas can be read in several different, semantically equivalent reading pathways (e.g., \verb|1/2| vs. \verb|\frac{1}{2}|).
Recently, some RL-based OCR models~\cite{wang2025infinity,chen2025logics,poznanski2025olmocr} meticulously develop composite reward mechanisms to optimize complex layout analysis and reading order reasoning.
However, these early attempts empirically construct RL datasets that could improve performance, lacking consideration for high-entropy formatted text and proper RL design.

In this paper, we propose a novel Format Decoupled Reinforcement Learning (FD-RL) method for document OCR, employing a two-stage SFT-then-RL training paradigm and customized data engineering.
First, the customized data engineering provides a large-scale format-rich training corpus by integrating diverse data sources, including open-source datasets, real-world PDF documents, and synthetic OCR samples.
Then, the SFT stage learns powerful OCR reading capabilities from  diverse data, providing the foundation for RL reasoning.
Finally, the RL stage provides two key innovations for reasoning over document OCR, termed as entropy-based data filtration and format decoupled reward. The data filtration strategy aims to select high entropy document samples by ranking them according to their averaged token entropy, inferred by the SFT model. The format decoupled reward first carefully separates the content of different formats from the model output, and then tailors a unique reward function for each format.

The main contributions of this study are as follows:

\begin{itemize}

\item We propose format decoupled reinforcement learning, which explores a SFT-then-RL training paradigm for reasoning-after-reading over documents.

\item We construct a large-scale, format-rich corpus from diverse sources with optimized mixing ratios, establishing a robust SFT baseline for subsequent RL training.

\item We devise entropy-based data curation that identifies high-entropy document samples, and calculate format decoupled rewards with content-specific reward functions.

\item We achieve 90.41 on OmniDocBench, a SOTA performance over previous end-to-end methods. Comprehensive ablation studies, encompassing data, training, data filtering, and format decoupled reward, thoroughly validates our effectiveness.

\end{itemize}

%% file: sec/2_related_work.tex
\section{Related Work}
\label{sec:formatting}

\subsection{Traditional Pipeline-based OCR Models} 
Pipeline-based OCR models~\cite{wang2024mineru,cui2025paddleocr,Paruchuri2025Marker} decompose document parsing into sequential stages: initial layout analysis to localize and segment content regions (including text, formulas, and tables), followed by region-specific parsers that extract and sequence outputs in reading order. For instance, MinerU~\cite{wang2024mineru} employs PDF-Extract-Kit~\cite{OpenDataLab2025} coupled with refined pre- and post-processing strategies to ensure extraction accuracy across varied document formats, while PP-StructureV3~\cite{cui2025paddleocr} combines layout detection, table recognition, and structural analysis in a multi-component pipeline to enable comprehensive document parsing. Although these pipeline-based models ensure stability through specialized components, they exhibit constrained adaptability and require significant effort for domain-specific customization.

\subsection{VLM-based OCR Methods} 

\subsubsection{End-to-End VLMs}
End-to-end VLMs~\cite{ocrflux2025,mistral2025ocr,liu2025points,poznanski2025olmocr-1,mandalm2025nanonets,wei2025deepseek,rednote2025dotsocr} adopt unified architectures that directly generate structured outputs from document images in a single forward pass, enabling streamlined processing and robust cross-domain generalization. Recent work in this direction includes POINTS-Reader~\cite{liu2025points}, which leverages public vision-language backbones with a training paradigm that synthesizes high-quality data and employs self-improvement strategies to bridge the gap between synthetic and real-world distributions. Similarly, dots.ocr~\cite{rednote2025dotsocr} utilizes native-resolution vision encoders to process documents at full fidelity while curating large-scale document corpora to enhance model performance. These methods demonstrate the effectiveness of unified VLM architectures for document parsing tasks.

\begin{figure*}[!t]
\centering
\includegraphics[width=\textwidth]{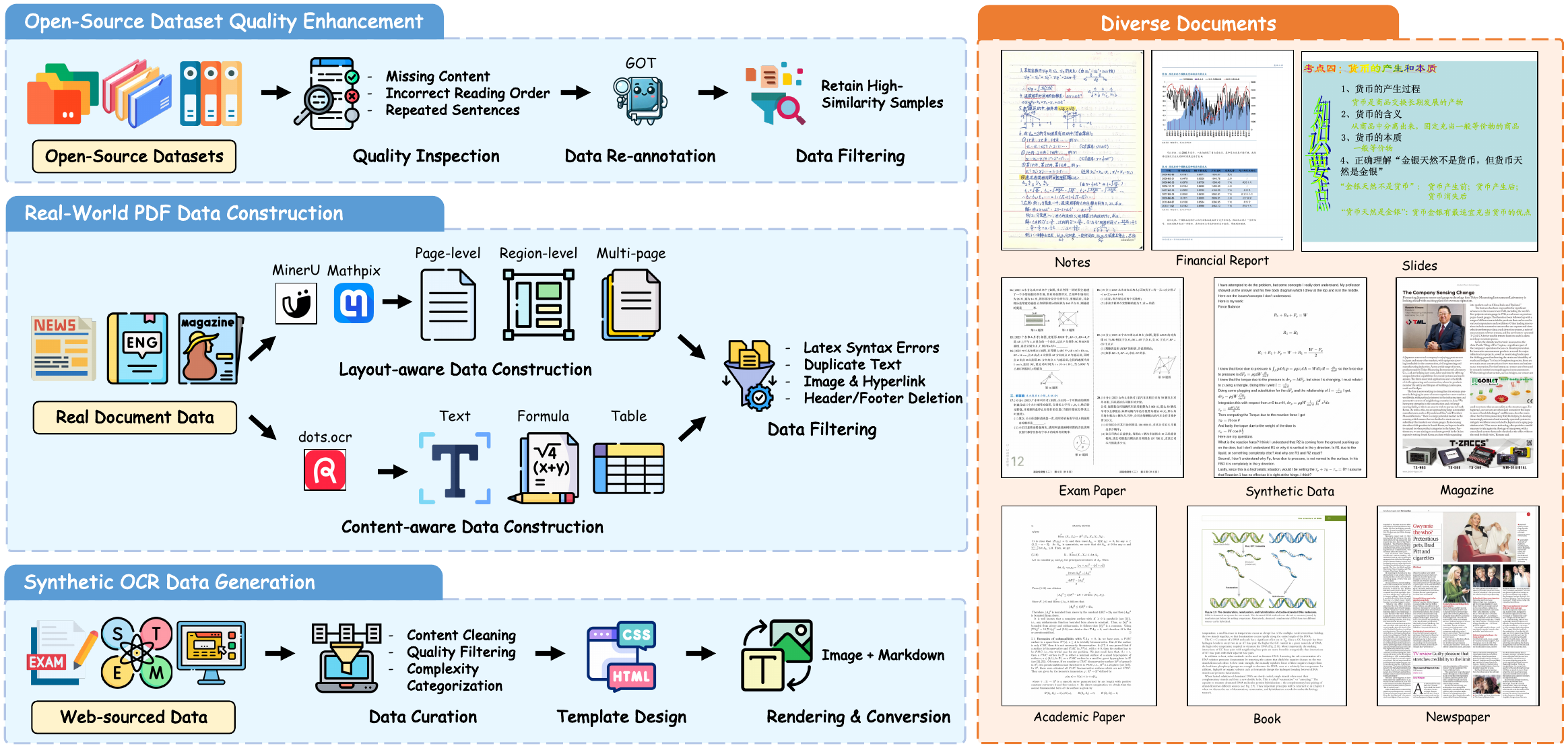}
\caption{Overview of our training data construction pipeline. We construct the dataset via (1) open-source dataset quality enhancement, (2) real-world PDF construction, and (3) synthetic OCR data generation, yielding a large-scale dataset covering nine common document categories: notes, financial reports, slides, exam papers, synthetic data, magazines, academic papers, books, and newspapers.}
\label{fig:data_all}
\end{figure*}

\subsubsection{Pipeline-based VLMs}
Pipeline-based VLMs~\cite{feng2025dolphin,li2025monkeyocr,niu2025mineru2,cui2025paddleocr-vl} decompose document parsing into sequential stages while leveraging VLMs for enhanced representation learning. MinerU2.5~\cite{niu2025mineru2} adopts a hierarchical parsing strategy that combines efficient layout analysis on downsampled inputs with fine-grained content extraction on high-resolution crops. In contrast, PaddleOCR-VL~\cite{cui2025paddleocr-vl} integrates traditional OCR for layout detection with a unified VLM for holistic content understanding. By combining structured layout analysis with VLM-based semantic reasoning, these methods achieve competitive performance while maintaining computational efficiency.

%% file: sec/3_method.tex
\section{Methodology}
\label{sec:method}
In this section, we first present systematic data engineering to construct a comprehensive training corpus. Then, we introduce a two-stage SFT-then-RL training strategy to achieve format decoupled optimization.

\subsection{Dataset Engineering}
\label{sec:data}
As illustrated in Figure~\ref{fig:data_all}, we construct a comprehensive training dataset through three methods: (1) open-source dataset quality enhancement, (2) real-world pdf data construction, and (3) synthetic OCR data generation.

\subsubsection{Open-Source Dataset Quality Enhancement} 
Open-source datasets offer low acquisition costs and large-scale coverage but suffer from quality inconsistencies due to heterogeneous sources and inconsistent annotation standards. We address this through a systematic enhancement pipeline. 

First, we collect diverse open-source document OCR datasets, including PDFA~\cite{pdfa}, DocStruct4M~\cite{hu2024mplug}, and DocGenome~\cite{xia2024docgenome} for office documents, forms, and structured text. We further supplement these with handwritten text and formula datasets such as IAM~\cite{marti2002iam}, ORAND-CAR~\cite{diem2014icfhr}, and HME~\cite{yuan2022syntax}, ensuring comprehensive scenario coverage. Second, we conduct quality inspection through random sampling and discover prevalent issues across the collected datasets: missing content, incorrect reading order, and repeated sentences. To address these issues, we employ the lightweight GOT~\cite{wei2024general} model to re-annotate all collected data, delivering high-quality annotations while maintaining computational efficiency. Finally, we apply similarity-based filtering to ensure data quality. We compute similarity between original labels and VLM-generated labels, retaining only samples with high similarity scores, which indicates high-quality original annotations. Importantly, we preserve the original labels as ground truth rather than VLM outputs to prevent overfitting to VLM-specific patterns.

\subsubsection{Real-World PDF Data Construction} 
Despite their advantages, open-source datasets remain insufficient in covering diverse real-world scenarios. To address this, we collect real-world PDF documents with diverse layout arrangements, font styles, color schemes, and varying resolutions, then perform precise annotation.
We categorize PDF processing into two types based on training objectives: layout-aware construction for structural understanding and content-aware construction for textual recognition.

\begin{figure*}[!t]
\centering
\includegraphics[width=\textwidth]{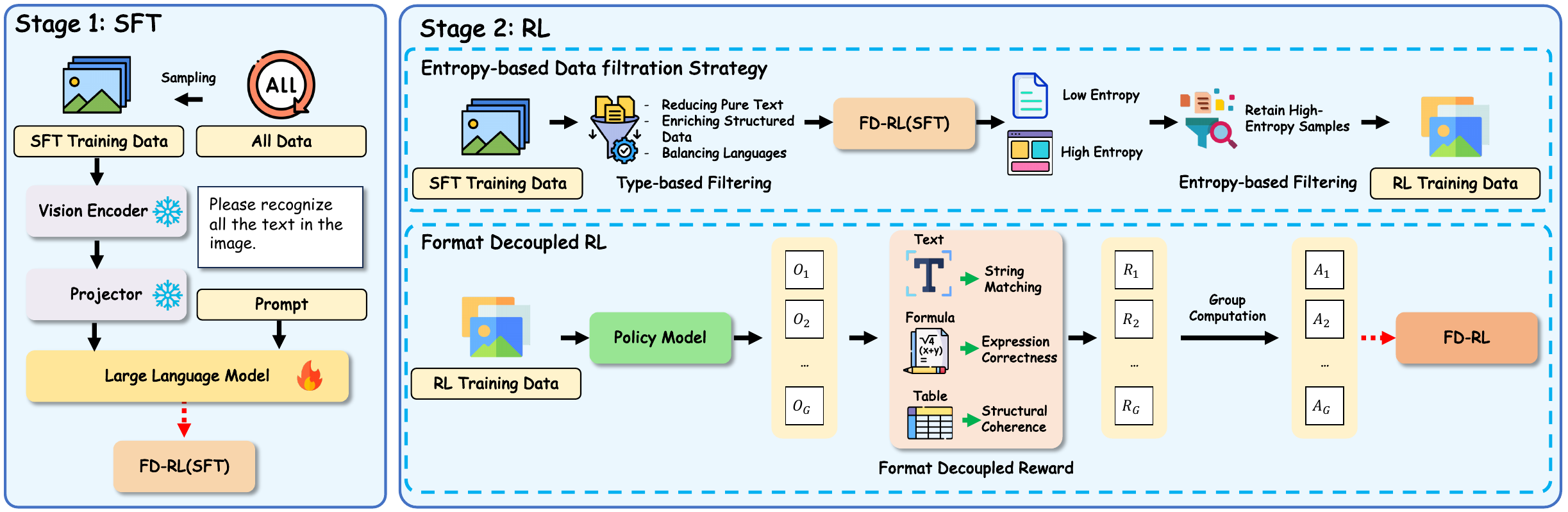}
\caption{\textbf{Overview of the FD-RL training pipeline.} Our approach comprises two stages: \textbf{Stage 1 (SFT)} trains FD-RL(SFT) on large-scale document data via supervised fine-tuning, freezing the vision encoder and projector while updating the language model. \textbf{Stage 2 (RL)} introduces two innovations: (1) \textit{Entropy-based Data Filtration Strategy} that filters data through type-based filtering (enriching structured data and balancing languages) and entropy-based filtering (retaining high-entropy samples), and (2) \textit{Format Decoupled RL} using separately for different content types: string matching reward for plain text, expression correctness reward for formulas, and structural coherence reward for tables.}
\label{fig:model_pipeline}
\end{figure*}

\paragraph{Layout-Aware Data Construction.} 
We collect a large and diverse set of documents and construct three progressive granularity levels to enhance multi-scale layout understanding. \textit{(1) Page-level data.} We employ MinerU~\cite{wang2024mineru} for OCR recognition to generate Markdown-format labels, validate LaTeX syntax using Mathjax, and eliminate duplicates via Text-Dedup~\cite{chenghao_mou_2023_8364980}. \textit{(2) Region-level data.} Following Fox~\cite{liu2024focus}'s approach, we use Mathpix to obtain coordinate-aware OCR results, then randomly select regions and aggregate their text lines into region boxes. We create color-guided annotations by marking boxes in different colors and synthesize negative samples with empty boxes to improve robustness. \textit{(3) Multi-page data.} We sample from page-level and region-level data to construct 2-6 page sequences, enabling cross-page parsing capability for complex document scenarios.

\paragraph{Content-Aware Data Construction.} 
Targeting accurate recognition of complete document text (plain text, formulas, tables), we collect diverse PDFs to capture varied symbol systems, fonts, and table structures. We apply dots.ocr~\cite{rednote2025dotsocr} for recognition, which handles complex elements by converting formulas to LaTeX, tables to HTML, and rearranging layouts in reading order to produce Markdown output. Post-processing ensures quality through image link removal (retaining pure text) and header/footer deletion (eliminating interference), yielding high-quality structured annotations.

\subsubsection{Synthetic OCR Data Generation}
While real PDF data closely resembles real-world scenarios, educational data and academic data remains scarce due to high manual annotation costs that hinder large-scale labeled data creation. 

To address these gaps, we generate synthetic OCR data from online resources, achieving both scale and quality. 
We first collect two types of content: practice problems from educational platforms spanning K12 to adult levels across multiple subjects and question types, and {academic Q\&A pairs from StackExchange covering graduate STEM topics}. After cleaning, filtering, and categorizing the collected data, we design HTML templates with realistic visual parameters including fonts, spacing, and colors, incorporating MathJax for LaTeX rendering and CSS for table formatting. We then inject the processed content and use Playwright to render high-resolution images, which are paired with Markdown text to form complete training samples.


\subsection{Model Training}
\label{sec:training}
Building upon the comprehensive dataset, we adopt a two-stage training strategy. 
As illustrated in Figure~\ref{fig:model_pipeline}, we first establish a strong OCR model through SFT. We then employ a format decoupled RL framework, incorporating entropy-based data filration strategy and a format decoupled reward, to guide the model toward learning the underlying format rules rather than memorizing token sequences.

\begin{table*}[!ht]
\centering
\caption{Performance comparison on OmniDocBench. \textbf{Bold} and \underline{underline} denote the best and second-best performance among end-to-end specialized VLMs, respectively. RD: Release Date, E2E: End-to-end (\ding{51}) or pipeline-based (\ding{55}), RO: Reading Order.}
\label{tab:omnidocbench}
\resizebox{\textwidth}{!}{
\begin{tabular}{llccccccccc}
\toprule
\textbf{Model Type} & \textbf{Methods} & \textbf{RD} & \textbf{E2E} & \textbf{Overall}$\uparrow$ & \textbf{Text}$^{\text{Edit}}\downarrow$ & \textbf{Formula}$^{\text{CDM}}\uparrow$ & \textbf{Table}$^{\text{TEDS}}\uparrow$ & \textbf{Table}$^{\text{TEDS-S}}\uparrow$ & \textbf{RO}$^{\text{Edit}}\downarrow$ \\
\midrule
\multirow{3}{*}{Pipeline Tools} 
& Marker-1.8.2~\cite{Paruchuri2025Marker} & 2025 & \ding{55} & 71.30 & 0.206 & 76.66 & 57.88 & 71.17 & 0.250 \\
& Mineru2-pipeline~\cite{wang2024mineru} & 2025 & \ding{55}  & 75.51 & 0.209 & 76.55 & 70.90 & 79.11 & 0.225 \\
& PP-StructureV3~\cite{cui2025paddleocr} & 2025 & \ding{55}  & 86.73 & 0.073 & 85.79 & 81.68 & 89.48 & 0.073 \\
\midrule
\multirow{5}{*}{General VLMs} 
& GPT-4o~\cite{achiam2023gpt} & 2023 & \ding{51} & 75.02 & 0.217 & 79.70 & 67.07 & 76.09 & 0.148 \\
& InternVL3-76B~\cite{zhu2025internvl3} & 2025 & \ding{51}  & 80.33 & 0.131 & 83.42 & 70.64 & 77.74 & 0.113 \\
& InternVL3.5-241B~\cite{wang2025internvl3} & 2025 & \ding{51}  & 82.67 & 0.142 & 87.23 & 75.00 & 81.28 & 0.125 \\
& Qwen2.5-VL-72B~\cite{Qwen2.5-VL} & 2025 & \ding{51}  & 87.02 & 0.094 & 88.27 & 82.15 & 86.22 & 0.102 \\
& Gemini-2.5 Pro~\cite{comanici2025gemini} & 2025  & \ding{51}  & 88.03 & 0.075 & 85.82 & 85.71 & 90.29 & 0.097 \\
\midrule
\multirow{15}{*}{Specialized VLMs} 
& Dolphin~\cite{feng2025dolphin} & 2025.05 & \ding{55} & 74.67 & 0.125 & 67.85 & 68.70 & 77.77 & 0.124 \\
& MinerU2-VLM~\cite{wang2024mineru} & 2025.06 & \ding{55} & 85.56 & 0.078 & 80.95 & 83.54 & 87.66 & 0.086 \\
& MonkeyOCR-pro-1.2B~\cite{li2025monkeyocr} & 2025.07 & \ding{55} & 86.96 & 0.084 & 85.02 & 84.24 & 89.02 & 0.130 \\
& MonkeyOCR-3B~\cite{li2025monkeyocr} & 2025.06 & \ding{55} & 87.13 & 0.075 & 87.45 & 81.39 & 85.92 & 0.129 \\
& MonkeyOCR-pro-3B~\cite{li2025monkeyocr} & 2025.07 & \ding{55} & 88.85 & 0.075 & 87.25 & 86.78 & 90.63 & 0.128 \\
& MinerU2.5~\cite{niu2025mineru2} & 2025.09 & \ding{55} & 90.67 & 0.047 & 88.46 & 88.22 & 92.38 & 0.044 \\
& PaddleOCR-VL~\cite{cui2025paddleocr-vl} & 2025.10 & \ding{55} & 92.56 & 0.035 & 91.43 & 89.76 & 93.52 & 0.043 \\
\cmidrule{2-10}
& OCRFlux-3B~\cite{ocrflux2025} & 2025.06 & \ding{51} & 74.82 & 0.193 & 68.03 & 75.75 & 80.23 & 0.202 \\
& Mistral OCR~\cite{mistral2025ocr} & 2025.03 & \ding{51} & 78.83 & 0.164 & 82.84 & 70.03 & 78.04 & 0.144 \\
& POINTS-Reader~\cite{liu2025points} & 2025.08 & \ding{51} & 80.98 & 0.134 & 79.20 & 77.13 & 81.66 & 0.145 \\
& olmOCR-7B~\cite{poznanski2025olmocr-1} & 2025.02 & \ding{51} & 81.79 & 0.096 & \underline{86.04} & 68.92 & 74.77 & 0.121 \\
& Nanonets-OCR-s~\cite{mandalm2025nanonets} & 2025.06 & \ding{51} & 85.59 & 0.093 & 85.90 & 80.14 & 85.57 & 0.108 \\
& Deepseek-OCR~\cite{wei2025deepseek} & 2025.10 & \ding{51} & 87.01 & 0.073 & 83.37 & {84.97} & {88.80} & 0.086 \\
& dots.ocr~\cite{rednote2025dotsocr} & 2025.07 & \ding{51} & \underline{88.41} & \textbf{0.048} & 83.22 & \underline{86.78} & \underline{90.62} & \textbf{0.053} \\
& FD-RL & 2025.11 & \ding{51} & \textbf{90.41} & \underline{0.049} & \textbf{88.67} & \textbf{87.35} & \textbf{92.10} & \underline{0.055} \\
\bottomrule
\end{tabular}
}
\end{table*}

\subsubsection{SFT Stage}
\label{sec:sft_stage}
To construct a domain-specialized OCR system, we first perform SFT on the pre-trained VLM (Qwen3-VL-4B~\cite{Qwen3-VL}) using large-scale datasets that encompass a wide range of document types, layouts, and formats. During SFT, we freeze the visual encoder (ViT) and multi-layer perceptron (MLP) modules, while updating only the parameters of the large language model (LLM). This strategy not only leverages the strong visual representations already learned by the backbone, but also concentrates computational resources on improving sequence decoding and format generation. To effectively utilize the training data, we progressively sample data from three sources, ultimately achieving a high-performance OCR model through SFT, thereby establishing a solid foundation for subsequent reinforcement learning.

\subsubsection{RL Stage}
Although SFT establishes a strong baseline for OCR tasks, it faces significant challenges in handling format-intensive content such as formulas and tables. To address these limitations, we introduce format decoupled RL as a complementary training paradigm that explicitly optimizes for format correctness and structural validity. Unlike SFT's token-level objective, format decoupled RL enables the model to receive targeted feedback on format-specific errors through carefully designed reward signals. For example, structural mistakes such as malformed LaTeX syntax or broken table hierarchies are given priority during optimization. This approach effectively mitigates the issue of format errors being drowned out by content errors, allowing the model to learn robust format-level representations that are critical for practical OCR applications.


\paragraph{Entropy-based Data Filtration.} 
To collect the RL dataset, we first select a subset with a high proportion of formulas and tables from the existing SFT data pool and balance the ratio of Chinese and English samples. To ensure data diversity and effectiveness, we remove samples that only contain plain text. Subsequently, we adopt an entropy-based data filtration strategy to further screen candidate data. Specifically, we use the SFT-trained model to perform inference on each sample, obtain the log probability (logprobs) of each token generated by the model, and calculate the average output entropy for each sample using the following formula:
\begin{equation}
\mathcal{D}_{\text{filtered}} = \left\{ d \in \mathcal{D}_{\text{raw}}  \middle| -\frac{1}{N_{d}} \sum_{i=1}^{N_{d}} \log {p_i}^{(d)} \geq \tau \right\}
\end{equation}
where $\tau$ is the entropy threshold, each $d$ is a sample from the original dataset, $N_d$ is the token number of $d$, and $p_i^{(d)}$ denotes the probability of the $i$-th token in $d$.
This filtration process systematically selects samples with greater structural complexity and prediction uncertainty, resulting in a challenging and representative dataset. Concentrating subsequent RL training on these high-entropy instances strengthens the model’s capacity for diverse reasoning and improves robustness on format-intensive content.

\begin{table*}[!t]
\centering
\caption{Text edit distance across different document types on OmniDocBench. \textbf{Bold} and \underline{underline} denote the best and second-best performance among end-to-end specialized VLMs, respectively. RD: Release Date, E2E: End-to-end (\ding{51}) or pipeline-based (\ding{55})}
\label{tab:text_edit_distance}
\resizebox{\textwidth}{!}{
\begin{tabular}{llcccccccccccc}
\toprule
\textbf{Model Type} & \textbf{Methods} & \textbf{RD} & \textbf{E2E} & \textbf{Slides} & \textbf{Academic Papers} & \textbf{Book} & \textbf{Textbook} & \textbf{Exam Paper} & \textbf{Magazine} & \textbf{Newspaper} & \textbf{Notes} & \textbf{Financial Report} \\
\midrule
\multirow{3}{*}{Pipeline Tools} 
& Marker-1.8.2~\cite{Paruchuri2025Marker} & 2025 & \ding{55} & 0.1796 & 0.0412 & 0.1010 & 0.2908 & 0.2958 & 0.1111 & 0.2717 & 0.4656 & 0.0341 \\
& Mineru2-pipeline~\cite{wang2024mineru} & 2025 & \ding{55} & 0.4244 & 0.0230 & 0.2628 & 0.1224 & 0.0822 & 0.3950 & 0.0736 & 0.2603 & 0.0411 \\
& PP-StructureV3~\cite{cui2025paddleocr} & 2025 & \ding{55} & 0.0794 & 0.0236 & 0.0415 & 0.1107 & 0.0945 & 0.0722 & 0.0617 & 0.1236 & 0.0181 \\
\midrule
\multirow{5}{*}{General VLMs} 
& GPT-4o~\cite{achiam2023gpt} & 2023 & \ding{51} & 0.1019 & 0.1203 & 0.1288 & 0.1599 & 0.1939 & 0.1420 & 0.6254 & 0.2611 & 0.3343 \\
& InternVL3-76B~\cite{zhu2025internvl3} & 2025 & \ding{51} & 0.0349 & 0.1052 & 0.0629 & 0.0827 & 0.1007 & 0.0406 & 0.5826 & 0.0924 & 0.0665 \\
& InternVL3.5-241B~\cite{wang2025internvl3} & 2025 & \ding{51} & 0.0475 & 0.0857 & 0.0237 & 0.1061 & 0.0933 & 0.0577 & 0.6403 & 0.1357 & 0.1117 \\
& Qwen2.5-VL-72B~\cite{Qwen2.5-VL} & 2025 & \ding{51} & 0.0422 & 0.0801 & 0.0586 & 0.1146 & 0.0681 & 0.0964 & 0.2380 & 0.1232 & 0.0264 \\
& Gemini-2.5 Pro~\cite{comanici2025gemini} & 2025 & \ding{51} & 0.0326 & 0.0182 & 0.0694 & 0.1618 & 0.0937 & 0.0161 & 0.1347 & 0.1169 & 0.0169 \\
\midrule
\multirow{11}{*}{Specialized VLMs} 
& Dolphin~\cite{feng2025dolphin} & 2025.05 & \ding{55} & 0.0957 & 0.0453 & 0.0616 & 0.1333 & 0.1684 & 0.0702 & 0.2388 & 0.2561 & 0.0186 \\
& MinerU2-VLM~\cite{wang2024mineru} & 2025.06 & \ding{55} & 0.0745 & 0.0104 & 0.0357 & 0.1276 & 0.0698 & 0.0652 & 0.1831 & 0.0803 & 0.0236 \\
& MonkeyOCR-pro-1.2B~\cite{li2025monkeyocr} & 2025.07 & \ding{55} & 0.0961 & 0.0354 & 0.0530 & 0.1110 & 0.0887 & 0.0494 & 0.0995 & 0.1686 & 0.0198 \\
& MonkeyOCR-pro-3B~\cite{li2025monkeyocr} & 2025.07 & \ding{55} & 0.0904 & 0.0362 & 0.0489 & 0.1072 & 0.0745 & 0.0475 & 0.0962 & 0.1165 & 0.0196 \\
& MinerU2.5~\cite{niu2025mineru2} & 2025.09 & \ding{55} & 0.0294 & 0.0235 & 0.0332 & 0.0499 & 0.0681 & 0.0316 & 0.0540 & 0.1161 & 0.0104 \\
\cmidrule{2-13}
& OCRFlux-3B~\cite{ocrflux2025} & 2025.06 & \ding{51} & 0.0870 & 0.0867 & 0.0818 & 0.1843 & 0.2072 & 0.1048 & 0.7304 & 0.1567 & 0.0193 \\
& Mistral OCR~\cite{mistral2025ocr} & 2025.03 & \ding{51} & 0.0917 & 0.0531 & 0.0610 & 0.1341 & 0.1341 & 0.0581 & 0.5643 & 0.3097 & 0.0523 \\
& POINTS-Reader~\cite{liu2025points} & 2025.08 & \ding{51} & 0.0334 & 0.0779 & 0.0671 & 0.1372 & 0.1901 & 0.1343 & 0.3789 & \underline{0.0937} & 0.0951 \\
& olmOCR-7B~\cite{poznanski2025olmocr-1} & 2025.02& \ding{51} & 0.0497 & {0.0365} & 0.0539 & 0.1204 & 0.0728 & 0.0697 & 0.2916 & 0.1220 & 0.0459 \\
& Nanonets-OCR-s~\cite{mandalm2025nanonets} & 2025.06 & \ding{51} & 0.0551 & 0.0578 & 0.0606 & 0.0931 & 0.0834 & 0.0917 & 0.1965 & 0.1606 & 0.0395 \\
& dots.ocr~\cite{rednote2025dotsocr} & 2025.07 & \ding{51} & \underline{0.0290} & \textbf{0.0231} & \underline{0.0433} & \textbf{0.0788} & \underline{0.0467} & \textbf{0.0221} & \textbf{0.0667} & 0.1116 & \textbf{0.0076} \\
& FD-RL & 2025.11 & \ding{51} & \textbf{0.0235} & \underline{0.0258} & \textbf{0.0300} & \underline{0.0867} & \textbf{0.0464} & \underline{0.0235} & \underline{0.1069} & \textbf{0.0881} & \underline{0.0091} \\
\bottomrule
\end{tabular}
}
\end{table*}

\paragraph{Format Decoupled Reward.}
To better guide the model in learning to parse formulas and tables and prevent the model from overlooking them in documents with lengthy text, we calculate rewards separately for different content types: string matching reward for plain text, expression correctness reward for formulas, and structural coherence reward for tables.
Specifically, we use regular expressions to extract formulas, tables, and plain text from both the model output and the ground truth. 
For plain text, normalized edit distance is used for character-level supervision; for formulas, we first convert them to equivalent LaTeX expressions and then apply the BLEU score for correctness supervision; for tables, TEDS is used to supervise structural consistency. The overall format decoupled reward $R$ is defined as:
\begin{align}
R = \frac{1}{\sum_{c=1}^{C} \mathbb{I}[|{GT}_c| > 0]}
\sum_{c=1}^{C} \mathbb{I}[|{GT}_c| > 0] \cdot
    f_c(Pred_c, GT_c)
\end{align}
where $C$ is the total number of content types, $Pred_{c}$ and $GT_{c}$ denote the model output and ground truth for the $c$-th type, and $I[\cdot]$ is the indicator function, which is 1 if the ground truth for that type is not empty and 0 otherwise. The function $f_c(.,.)$ denotes the reward function specific to content type $c$.
This design enables fine-grained rewards for different content types in complex document structures, effectively improving the model’s ability to parse format-intensive content such as formulas and tables, and enhancing its robustness in handling diverse document layouts.

\noindent \textbf{Format-Decoupled RL.} 
During the RL stage, we fine-tune the model using GRPO~\cite{shao2024deepseekmath}, which optimizes policy directly with group-normalized rewards from sampled responses, without requiring a separate critic. For each input $x$, multiple responses $\{o_1,o_2,...,o_G\}$ are sampled from the current policy $\theta_\mathrm{old}$, and the reward $R_i$ are computed. The group-normalized advantage for the $i$-th response is:
\begin{equation}
{A}_{i} = \frac{R_i - \mathrm{mean}(\{R_j\}_{j=1}^G)}{\mathrm{std}(\{R_j\}_{j=1}^G)}
\end{equation}
The model parameters $\theta$ are then optimized by maximizing the following objective function:
\begin{align}
J_{\mathrm{GRPO}}(\theta) = & \mathbb{E}_{(x, y) \sim D, \{o_i\}_{i=1}^G \sim \pi_{\theta_\mathrm{old}}(\cdot|x)} \\
    \Bigg[ 
    & \frac{1}{G} \sum_{i=1}^G  
    \min \Big( 
        \frac{\pi_\theta(o_{i} | x)}{\pi_{\theta_\mathrm{old}}(o_{i} | x)} {A}_{i}, \nonumber \\
    & \mathrm{clip}(\frac{\pi_\theta(o_{i} | x)}{\pi_{\theta_\mathrm{old}}(o_{i} | x)}, 1-\epsilon, 1+\epsilon) {A}_{i}
    \Big) \notag
\Bigg]
\end{align}
where $\epsilon$ is the hyperparameter, $\pi_\theta$ and $\pi_{\theta_\mathrm{old}}$ are the optimized model and the policy model, respectively.

%% file: sec/4_experiments.tex
\section{Experiment}
\label{sec:experiments}
In this section, we present experimental settings, benchmark our method against state-of-the-art OCR models, and provide comprehensive analysis including qualitative results and ablation studies.

\subsection{Experimental Settings}
\noindent \textbf{Dataset.} 
OmniDocBench~\cite{ouyang2025omnidocbench} is a comprehensive document parsing benchmark containing 1,355 pages across nine diverse document types (books, notes, slides, etc.). It provides rich elements including text, formulas, and tables, covers two languages and varied layout structures.

\noindent \textbf{Metrics.} 
Following the official OmniDocBench  evaluation protocol, we employ three specialized metrics to assess different aspects of document parsing performance: Edit Distance for text recognition accuracy, CDM for formula recognition, and TEDS for table structure recognition. The overall document parsing score is computed as:

\begin{equation} \text{Overall} = \frac{(1 - \text{Text}^{\text{Edit}}) \times 100 + \text{Formula}^{\text{CDM}} + \text{Table}^{\text{TEDS}}}{3}. \end{equation}

\subsection{Quantitative Comparisons}
We compare FD-RL against pipeline tools~\cite{wang2024mineru,cui2025paddleocr,Paruchuri2025Marker}, general VLMs~\cite{achiam2023gpt,zhu2025internvl3,wang2025internvl3,Qwen2.5-VL,comanici2025gemini}, and specialized VLMs including end-to-end models~\cite{ocrflux2025,mistral2025ocr,liu2025points,poznanski2025olmocr-1,mandalm2025nanonets,wei2025deepseek,rednote2025dotsocr} and pipeline-based models~\cite{feng2025dolphin,li2025monkeyocr,niu2025mineru2,cui2025paddleocr-vl}. Evaluation covers (1) performance across content types and (2) performance across document types.


\subsubsection{Performance Across Content Types} 
As shown in Table~\ref{tab:omnidocbench}, FD-RL demonstrates superior performance on OmniDocBench with an overall score of 90.41, surpassing dots.ocr by 2.0 points and Deepseek-OCR by 3.4 points among end-to-end VLMs. For text recognition, FD-RL attains an edit distance of 0.049, ranking second only to dots.ocr (0.048) while substantially outperforming Deepseek-OCR (0.073). In formula recognition, FD-RL establishes the best CDM score of 88.67, exceeding olmOCR-7B (86.04) and Nanonets-OCR-s (85.90). For table parsing, FD-RL achieves the top TEDS score of 87.35 and TEDS-S score of 92.10. In reading order prediction, it secures the second-best edit distance of 0.055. These results demonstrate the effectiveness of our reinforcement learning framework in optimizing format-level validity across diverse document parsing tasks.

\subsubsection{Performance Across Document Types} 
Table~\ref{tab:text_edit_distance} presents document-type-specific results, where FD-RL secures the top performance in 4 out of 9 categories and ranks second in the remaining 5 categories. Specifically, FD-RL achieves the lowest edit distance for slides (0.0235) and exam papers (0.0464), outperforming all baseline methods. For books and notes, it attains edit distances of 0.0300 and 0.0881, respectively, significantly surpassing dots.ocr (0.0433) and POINTS-Reader (0.0937). In the remaining categories, FD-RL consistently ranks second with competitive scores that closely approach the best performance. This robust cross-domain generalization can be attributed to our comprehensive training corpus, which integrates large-scale open-source data, real-world PDF documents, and synthetic OCR samples, enabling the model to adapt effectively across diverse document types.

\begin{table}[!t]
\centering
\small
\caption{Ablation study on multi-source data. OSD: open-source datasets; RWPD: real-world PDF data; SOD: synthetic OCR data; RLD: reinforcement learning data; Table/Table$^{*}$: TEDS/TEDS-S.}
\label{tab:data_source}
\resizebox{\linewidth}{!}{
\begin{tabular}{ccc|c|cccccc}
\toprule
\textbf{OSD} & \textbf{RWPD} & \textbf{SOD} & \textbf{RLD} & \textbf{Overall}$\uparrow$ & \textbf{Text}$\downarrow$ & \textbf{Formula}$\uparrow$ & \textbf{Table}$\uparrow$ & \textbf{Table}$^{*}\uparrow$ & \textbf{RO}$\downarrow$ \\
\midrule
{\color[gray]{0.8} \ding{55}} & {\color[gray]{0.8} \ding{55}} & {\color[gray]{0.8} \ding{55}} & {\color[gray]{0.8} \ding{55}} & 46.06 & 0.553 & 55.40 & 38.06 & 42.18 & 0.371 \\
\ding{51} & {\color[gray]{0.8} \ding{55}} & {\color[gray]{0.8} \ding{55}} & {\color[gray]{0.8} \ding{55}} & 78.25 & 0.095 & 78.07 & 66.19 & 70.30 & 0.101 \\
\ding{51} & \ding{51} & {\color[gray]{0.8} \ding{55}} & {\color[gray]{0.8} \ding{55}} & 84.16 & 0.063 & 83.10 & 75.69 & 79.50 & 0.069 \\ 
\ding{51} & \ding{51} & \ding{51} & {\color[gray]{0.8} \ding{55}} & {87.13} & {0.055} & {85.60} & {81.27} & {84.91} & {0.063} \\
\ding{51} & \ding{51} & \ding{51} & \ding{51} & \textbf{90.41} & \textbf{0.049} & \textbf{88.67} & \textbf{87.35} & \textbf{92.10} & \textbf{0.055} \\
\bottomrule
\end{tabular}
}
\end{table}

\begin{table}[!t]
\centering
\small
\caption{Ablation study on the two-stage SFT-then-RL training strategy.}
\label{tab:stage}
\resizebox{\linewidth}{!}{
\begin{tabular}{cc|cccccc}
\toprule
\textbf{SFT} & \textbf{RL} & \textbf{Overall}$\uparrow$ & \textbf{Text}$\downarrow$ & \textbf{Formula}$\uparrow$ & \textbf{Table}$\uparrow$ & \textbf{Table}$^{*}\uparrow$ & \textbf{RO}$\downarrow$ \\
\midrule
{\color[gray]{0.8} \ding{55}} & {\color[gray]{0.8} \ding{55}} & 46.06 & 0.553 & 55.40 & 38.06 & 42.18 & 0.371 \\
{\color[gray]{0.8} \ding{55}} & \ding{51} & 49.37 & {0.460} & 62.03 & 32.08 & 34.32 & 0.319 \\
\ding{51} & {\color[gray]{0.8} \ding{55}} & 87.13 & 0.055 & 85.60 & 81.27 & 84.91 & 0.063 \\
\ding{51} & \ding{51} & \textbf{90.41} & \textbf{0.049} & \textbf{88.67} & \textbf{87.35} & \textbf{92.10} & \textbf{0.055} \\
\bottomrule
\end{tabular}
}
\end{table}

\begin{table}[!t]
\centering
\small
\caption{Ablation study on the entropy-based data filtration strategy.}
\label{tab:data_filter}
\resizebox{\linewidth}{!}{
\begin{tabular}{c|cccccc}
\toprule
\textbf{Filtration rate} & \textbf{Overall}$\uparrow$ & \textbf{Text}$\downarrow$ & \textbf{Formula}$\uparrow$ & \textbf{Table}$\uparrow$ & \textbf{Table}$^{*}\uparrow$ & \textbf{RO}$\downarrow$ \\
\midrule
0\% & 88.47 & 0.050 & 85.71 & 84.72 & 88.79 & 0.059 \\
25\% & 89.53 & 0.049 & 87.48 & 86.00 & 90.13 & 0.055 \\
50\% & \textbf{90.41} & \textbf{0.049} & \textbf{88.67} & \textbf{87.35} & \textbf{92.10} & \textbf{0.055} \\
75\% & 88.58 & 0.057 & 86.50 & 84.95 & 88.77 & 0.063 \\
\bottomrule
\end{tabular}
}
\end{table}

\begin{table}[!t]
\centering
\small
\caption{Ablation study on the format decoupled reward. FP: format separation over plain text, formula and table; SM/EC/SC: string match/expression correctness/structural coherence reward.}
\label{tab:reward}
\resizebox{\linewidth}{!}{
\begin{tabular}{c|ccc|cccccc}
\toprule
\textbf{FP} & \textbf{SM} & \textbf{EC} & \textbf{SC} & \textbf{Overall}$\uparrow$ & \textbf{Text}$\downarrow$ & \textbf{Formula}$\uparrow$ & \textbf{Table}$\uparrow$ & \textbf{Table}$^{*}\uparrow$ & \textbf{RO}$\downarrow$ \\
\midrule
{\textcolor{gray!80}{\ding{55}}} & \ding{51} & {\textcolor{gray!80}{\ding{55}}} & {\textcolor{gray!80}{\ding{55}}} & 88.64 & \textbf{0.044} & 87.04 & 83.27 & 87.44 & 0.058 \\
\midrule
\ding{51} &\ding{51} & {\textcolor{gray!80}{\ding{55}}} & {\textcolor{gray!80}{\ding{55}}} & 89.61 & 0.052 & 87.18 & 86.84 & 91.06 & 0.059 \\
\ding{51} &\ding{51} & \ding{51} & {\textcolor{gray!80}{\ding{55}}} & 89.80 & 0.050 & 87.59 & 86.80 & 90.71 & 0.057 \\
\ding{51} &\ding{51} & \ding{51} & \ding{51} & \textbf{90.41} & {0.049} & \textbf{88.67} & \textbf{87.35} & \textbf{92.10} & \textbf{0.055} \\
\bottomrule
\end{tabular}
}
\end{table}

\subsection{Ablation Studies}
\noindent \textbf{Multi-Source Data.} 
Table~\ref{tab:data_source} demonstrates the effectiveness of our multi-source data strategy. In the SFT stage, we start from a baseline without any training data, achieving an overall score of 46.06. Incorporating open-source datasets yields substantial improvement, reaching 78.25 (+32.19) and establishing foundational OCR capabilities. Adding real-world PDF data further enhances performance to 84.16 (+5.91), improving robustness to diverse document layouts and real-world variations. Integrating synthetic OCR data pushes the score to 87.13 (+2.97), providing diverse format-intensive samples that strengthen formula and table recognition. Finally, combining all SFT data with RL data achieves the highest overall score of 90.41 (+3.28), demonstrating the synergistic power of multi-source supervised learning and reinforcement learning. These results validate that each data source plays a complementary role, with their progressive integration being essential for achieving robust performance on format-intensive document parsing tasks.

\begin{figure*}[!t]
\centering
\includegraphics[width=\linewidth]{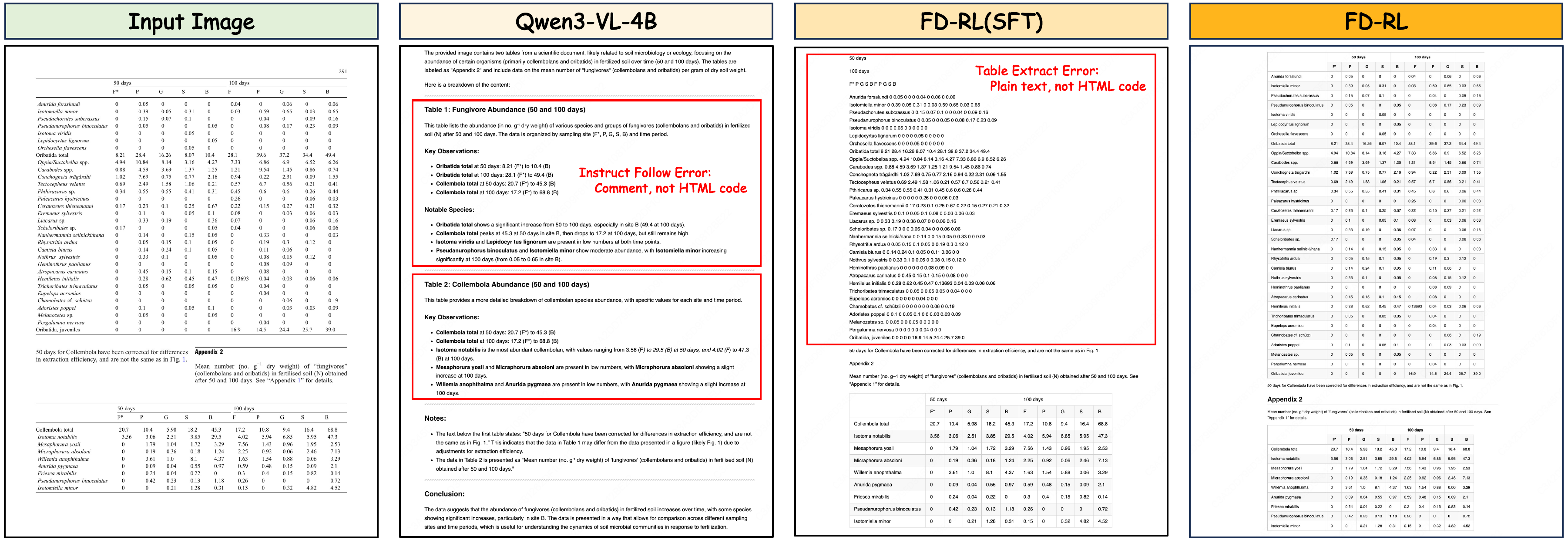}
\caption{Qualitative examples comparing FD-RL with baselines. FD-RL successfully follows instructions to generate HTML code and correctly extracts tables in structured format, demonstrating FD-RL's superiority in format-intensive content parsing.}
\vspace{-4mm}
\label{fig:doc-r1-visual}
\end{figure*}

\noindent \textbf{Two-Stage Training Strategy.} 
Table~\ref{tab:stage} demonstrates the necessity of our two-stage training approach. Applying RL directly to a general VLM without SFT yields limited improvement (49.37 overall, +3.31), as the base model lacks OCR-specific capabilities from fine-tuning on document data. In contrast, SFT alone achieves substantial gains (87.13, +41.07), validating the importance of domain adaptation through large-scale supervised learning. The full SFT+RL pipeline achieves the best performance (90.41), with RL providing additional improvements particularly in format-intensive tasks: formula (+3.07) and table (+6.08 TEDS, +7.19 TEDS-S). These results confirm that SFT establishes foundational OCR capabilities, while RL shifts optimization from token-level accuracy to format-level validity, enabling robust performance on format-intensive document parsing.

\noindent \textbf{Entropy-based Data Filtering.} 
Table~\ref{tab:data_filter} presents the ablation study on entropy-based data filtration, demonstrating the critical role of appropriate data filtration strategies. Without filtration (0\% filtration rate), the model achieves the lowest overall performance (88.47), indicating that entropy-based  data filtration is crucial as low-entropy samples reduce RL effectiveness on format-intensive content. Moreover, the filtration rate must be appropriately calibrated. While 50\% filtration rate achieves optimal performance (90.41), both 25\% (89.53) and 75\% (88.58) lead to performance degradation, suggesting that insufficient filtration fails to adequately concentrate on high-entropy instances, while excessive data filtration discards too many valuable samples. These results validate that entropy-based filtration with an appropriate rate is crucial for balancing sample quality and training diversity in format decoupled RL.

\noindent \textbf{Format Decoupled Reward.}
Table~\ref{tab:reward} demonstrates the necessity of our format decoupled reward design. Using unified string match (SM) reward without format separation achieves 88.64 overall, as it fails to account for distinct content characteristics. Introducing modality separation with SM improves performance to 89.61 (+0.97), but gains remain limited because SM operates purely at the token level, unable to distinguish format-level validity. Adding expression correctness (EC) rewards further enhances performance by evaluating content-specific accuracy, particularly benefiting formula recognition through semantic equivalence assessment. The full scheme with structural coherence (SC) rewards achieves the best results, as SC explicitly assesses format validity and structural integrity crucial for table parsing. These results validate that decomposing rewards into specialized components—each targeting specific aspects of format-level correctness—is essential for robust document parsing.

\subsection{Qualitative Visualizations}
Figure~\ref{fig:doc-r1-visual} presents qualitative examples comparing FD-RL with baselines on table extraction. The first column shows the input document containing two tables. The second column presents the result from general VLMs (Qwen3-VL-4B~\cite{Qwen3-VL}), which exhibits an instruction-following error by generating plain comments rather than the requested HTML code. The third column displays the output of FD-RL(SFT), which suffers from a table extraction error by producing unstructured plain text without preserving table structure. The fourth column demonstrates FD-RL's output, which successfully generates well-formatted HTML code that accurately captures the table structure and adheres to the instruction. These examples validate that FD-RL's format decoupled optimization effectively addresses both instruction-following and format-preservation challenges, enabling superior performance on format-intensive content parsing.

%% file: sec/5_conclusion.tex
\section{Conclusion}
We observe an intriguing phenomenon in document OCR: format-sensitive content exhibits significantly higher entropy than plain text due to diverse valid structural expressions, a perspective that has been largely overlooked in prior work. Following this insight, we propose format decoupled reinforcement learning (FD-RL), which leverages entropy-based filtration strategy and format decoupled rewards. We achieve an overall score of 90.41 on OmniDocBench, demonstrating notable gains on the complex document parsing task, thus validating the effectiveness of our approach.


%% file: sec/X_suppl.tex
\clearpage
\setcounter{page}{1}
\maketitlesupplementary

\section{Implementation Details}
In this section, we elaborate on the implementation details of FD-RL. We describe the data construction for both SFT and RL stages, followed by the training configurations and hyperparameters used in each stage.

\subsection{Training Data}
\subsubsection{SFT Data}
\label{sec:sft_data}

Our SFT training dataset comprises 566k samples from three complementary sources, as shown in Table~\ref{tab:data_samples}. Specifically, we collect 240k samples from open-source datasets, 208k samples from real-world PDF documents, and 118k synthetically generated OCR samples. This diverse composition ensures comprehensive coverage of various document types and OCR scenarios.

\begin{table}[!h]
\centering
\caption{Distribution of training samples across data sources}
\label{tab:data_samples}
\vspace{-2mm}
\begin{tabular}{l|r}
\toprule
\textbf{Source} & \textbf{Samples} \\ 
\midrule
Open-Source Dataset & 240k \\
Real-World PDF Data & 208k \\
Synthetic OCR Data & 118k \\
\midrule
\textbf{Total} & \textbf{566k} \\
\bottomrule
\end{tabular}
\vspace{-3mm}
\end{table}

\begin{table}[!h]
\centering
\caption{Distribution of training samples across document categories}
\label{tab:pdf_samples}
\vspace{-2mm}
\begin{tabular}{l|r}
\toprule
\textbf{Document Category} & \textbf{Samples} \\ 
\midrule
Notes & 6k \\ 
Financial Reports & 3k \\ 
Slides & 10k \\ 
Exam Papers & 3k \\ 
Synthetic Data & 118k \\ 
Magazines & 100k \\ 
Academic Papers & 136k \\ 
Books & 108k \\ 
Newspapers & 82k \\ 
\midrule
\textbf{Total} & \textbf{566k} \\
\bottomrule
\end{tabular}
\vspace{-3mm}
\end{table}

As detailed in Table~\ref{tab:pdf_samples}, the dataset spans nine distinct document categories, each representing common real-world OCR challenges. Academic papers constitute the largest portion with 136k samples, followed by books (108k) and magazines (100k), which collectively provide rich training data for long-form document understanding. Newspapers contribute 82k samples, capturing diverse layout patterns and multi-column structures. The synthetic data category includes 118k samples designed to complement educational materials that are difficult to obtain in real-world scenarios. Additionally, we incorporate slides (10k), notes (6k), financial reports (3k), and exam papers (3k) to ensure coverage of specialized document types with unique structural characteristics. This category-wise distribution enables the model to learn format-specific patterns while maintaining generalization across diverse document scenarios.

\subsubsection{RL Data}
To ensure the acquisition of format-intensive data for RL training, we manually remove document types that are primarily plain text, such as slides, magazines, books, and newspapers. Among the remaining data, we further filter out plain text samples that do not contain tables or formulas using regular expressions, resulting in a format-intensive candidate dataset of 16k samples. We then use the SFT-trained model to perform inference on all the data. For each sample, we calculate the average output entropy and retain the top 50\% of high-entropy samples for RL training. Our ablation studies demonstrate that this proportion is optimal.

\subsection{Training Details}

\subsubsection{SFT}
We fine-tune Qwen3-VL-4B~\cite{Qwen3-VL} into FD-RL(SFT) using the LLaMA Factory framework~\cite{zheng2024llamafactory} on two nodes with 8 GPUs each. The training employs the dataset from Section~\ref{sec:sft_data} with the Qwen3-VL template (without thinking tokens) and a maximum sequence length of 8,192 tokens. For efficient data processing, we use 128 preprocessing workers (batch size 256) and 128 dataloader workers. We adopt a learning rate of 1e-5 with cosine scheduling, 10\% warmup ratio, per-device batch size of 1, and train for one epoch using BF16 mixed precision. Following Section~\ref{sec:sft_stage}, we freeze the ViT and MLP modules while applying full parameter fine-tuning to the LLM, enabling comprehensive adaptation to OCR tasks.


\begin{figure*}[!t]
\centering
\includegraphics[width=0.75\linewidth]{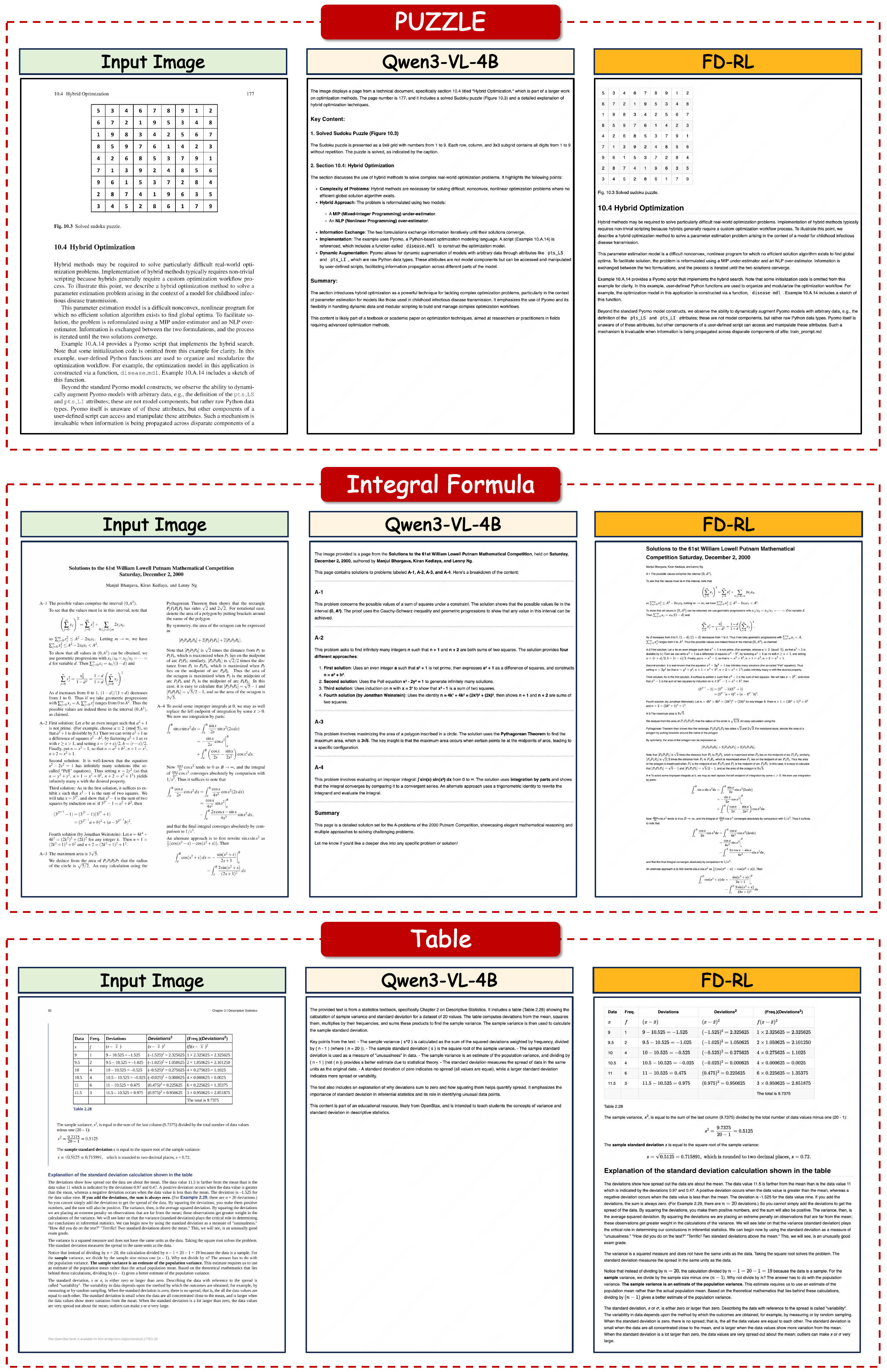}
\caption{Qualitative comparison of Qwen3-VL-4B and FD-RL on three types of formatted content: Puzzle, Integral Formula, and Table.}
\vspace{-4mm}
\label{fig:fd-rl-visual_supp}
\end{figure*}

\subsubsection{RL}
For the RL stage, we perform reinforcement learning on the FD-RL(SFT) model obtained from the SFT stage. Specifically, we adopt the EasyR1~\cite{zheng2025easyr1} framework built upon the veRL project for reinforcement learning training. In terms of batch configuration, both the rollout batch size and training batch size are set to 32. During the generation process, we use a temperature of 1 and generate 8 rollouts per sample, with the maximum response length capped at 10,240 tokens. Notably, KL divergence is excluded from the loss calculation to allow more flexible policy updates. Regarding optimization settings, we employ BF16 mixed precision training with a learning rate of 1e-6 and weight decay of 1e-2, while keeping all model parameters unfrozen throughout the training process. Finally, the complete RL dataset is trained for one epoch on a single node equipped with 8 GPUs, requiring approximately 10 hours to complete.

\section{Experiment Results}
This section validates FD-RL through quantitative comparison with RL-based OCR models and qualitative analysis of parsing results.

\subsection{Comparison with RL-based OCR models}

Table~\ref{tab:other_model} compares FD-RL with other RL-based OCR models. Infinity Parser~\cite{wang2025infinity} achieves an overall score of 56.12, showing limited capability in handling complex documents. olmOCR 2~\cite{poznanski2025olmocr} demonstrates substantial improvement with a score of 80.03, while Logics-Parsing~\cite{chen2025logics} further advances performance to 83.85, representing the previous state-of-the-art among RL-based approaches. Our FD-RL method achieves the best performance with an overall score of 90.41, surpassing the previous best by 6.56 points. This significant improvement demonstrates the effectiveness of our format-decoupled reinforcement learning approach in learning underlying format rules rather than memorizing token sequences.

\begin{table}[!ht]
\centering
\small
\caption{Comparison with RL-based OCR models.}
\label{tab:other_model}
\resizebox{\linewidth}{!}{
\begin{tabular}{c|cccccc}
\toprule
\textbf{Methods} & \textbf{Overall}$\uparrow$ & \textbf{Text}$\downarrow$ & \textbf{Formula}$\uparrow$ & \textbf{Table}$\uparrow$ & \textbf{Table}$^{*}\uparrow$ & \textbf{RO}$\downarrow$ \\
\midrule
Infinity Parser~\cite{wang2025infinity} & 56.12 & 0.494 & 73.70 & 44.05 & 47.48 & 0.304 \\
olmOCR 2~\cite{poznanski2025olmocr} & 80.03 & 0.163 & 84.00 & 72.40 & 76.60 & 0.145 \\
Logics-Parsing~\cite{chen2025logics} & 83.85 & 0.158 & 84.36 & 82.97 & 87.76 & 0.108 \\
FD-RL & \textbf{90.41} & \textbf{0.049} & \textbf{88.67} & \textbf{87.35} & \textbf{92.10} & \textbf{0.055} \\
\bottomrule
\end{tabular}
}
\end{table}

\subsection{Qualitative Results}

Figure~\ref{fig:fd-rl-visual_supp} compares Qwen3-VL-4B~\cite{Qwen3-VL} and FD-RL on three types of formatted content: Puzzle, Integral Formula, and Table.
{For Puzzle content}, FD-RL accurately recognizes the structure and content of mathematical puzzles, demonstrating superior performance in understanding complex problem layouts.
{In Integral Formula recognition}, FD-RL shows significant advantages in handling complex \LaTeX{} expressions with multi-line alignments and nested symbols. The model maintains better structural integrity when rendering integration formulas.
{For Table processing}, FD-RL accurately preserves both table structures and embedded mathematical notation, which is crucial for scientific document processing.
Overall, FD-RL consistently outperforms Qwen3-VL-4B across all three content types, demonstrating that reinforcement learning effectively enhances formatted content understanding and generation.